\documentclass[conference]{IEEEtran}
\IEEEoverridecommandlockouts
\usepackage{cite}
\usepackage{amsmath,amssymb,amsfonts}
\usepackage{algorithmic}
\usepackage{graphicx}
\usepackage{textcomp}
\usepackage{xcolor}
\usepackage{hyperref}
\usepackage[all]{hypcap}
\usepackage{soul}
\def\BibTeX{{\rm B\kern-.05em{\sc i\kern-.025em b}\kern-.08em
    T\kern-.1667em\lower.7ex\hbox{E}\kern-.125emX}}

\hypersetup{
  pdftitle={AI-Driven Decision-Making System for Hiring Process},
  pdfauthor={Vira Filatova; Andrii Zelenchuk; Dmytro Filatov},
  pdfsubject={Conference paper},
  pdfkeywords={AI-assisted hiring, candidate screening, decision support systems, large language models (LLMs), multi-agent systems, multimodal recruitment analytics, explainable scoring, human-in-the-loop validation, public-data verification, resume parsing and skill matching.}
}

\begin{document}

\title{AI-Driven Decision-Making System for Hiring Process}

\author
{\IEEEauthorblockN{Vira Filatova}
\IEEEauthorblockA{\textit{Applied Artificial Intelligence} \\
\textit{Covijn Ltd.}\\
London, United Kingdom \\
vira@deepxhub.com}
\and
\IEEEauthorblockN{Andrii Zelenchuk}
\IEEEauthorblockA{\textit{Artificial Intelligence and Computer Vision} \\
\textit{Aimech Technologies Corp.}\\
Kyiv, Ukraine \\
andrii.zelenchuk@deepxhub.com}
\and
\IEEEauthorblockN{Dmytro Filatov }
\IEEEauthorblockA{\textit{Artificial Intelligence and Computer Vision} \\
\textit{Aimech Technologies Corp.}\\
San Francisco, The United States of America \\
dima@deepxhub.com}
}

\maketitle

\begin{abstract}
Early-stage candidate validation is a major bottleneck in hiring, because recruiters must reconcile heterogeneous inputs (resumes, screening answers, code assignments, and limited public evidence). This paper presents an AI-driven, modular multi-agent hiring assistant that integrates (i) document and video preprocessing, (ii) structured candidate profile construction, (iii) public-data verification, (iv) technical/culture-fit scoring with explicit risk penalties, and (v) human-in-the-loop validation via an interactive interface. The pipeline is orchestrated by an LLM under strict constraints to reduce output variability and to generate traceable component-level rationales. Candidate ranking is computed by a configurable aggregation of technical fit, culture fit, and normalized risk penalties. The system is evaluated on 64 real applicants for a mid-level Python backend engineer role, using an experienced recruiter as the reference baseline and a second, less experienced recruiter for additional comparison. Alongside precision/recall, we propose an efficiency metric measuring expected time per qualified candidate. In this study, the system improves throughput and achieves 1.70 hours per qualified candidate versus 3.33 hours for the experienced recruiter, with substantially lower estimated screening cost, while preserving a human decision-maker as the final authority.
\end{abstract}


\section{Introduction}
Modern hiring process involves reviewing hundreds of applications for a single position while trying to follow a consistent, fair, and thorough evaluation process. Even for smaller companies, thoroughly reading resumes, conducting preliminary interviews, and checking candidates' online activity can easily take several days of a recruiter's or hiring manager's time for each position. At the same time, expectations for transparency and reduced bias in AI-based hiring tools are growing. Systems are now expected not only to save time, but also to maintain the quality of decisions at least as good as that of experienced humans, and to justify their recommendations in a way that practitioners can verify.

Most existing automation systems focus on individual steps in this workflow. Applicant tracking systems (ATS) and commercial resume screening tools use keyword filters or static machine learning models to rank resumes. Some products handle chatbot-style pre-screening, one-on-one video interviews, or social media checks. These tools can be effective in their respective niches, but they rarely offer a single, multi-modal view of a candidate based on all the information. As a result, hiring teams are often forced to combine multiple systems and manual checks, with limited control over how signals from different sources are combined and no clear idea of how much time they are actually saving compared to a thorough human review.

Our proposed system is based on a specific hiring scenario: a small, tech-savvy company is looking for a mid-level engineer with experience in Python backend. In this situation, the most time-consuming and error-prone part of the hiring process is filtering and selecting candidates after they have applied. This step involves basic resume screening, including compliance with basic technology, a short structured interview, and public record checks, as well as simple cues from the candidate, such as fit with corporate culture, tone, professionalism, and obvious red flags in the environment. Our goal is to automate this stage of the selection process as much as possible, leaving the final decision to the hiring manager. 

In response to these gaps, we are implementing and studying an AI-based candidate screening system that intentionally integrates multiple modalities and stages of reasoning into one coherent system. 

The main research question is pragmatic: how much time can such a system save compared to manual screening by an experienced hiring professional, while maintaining comparable quality of decisions? To investigate this, we evaluate the system on a real database with 64 candidates for a backend engineer position. For each candidate, we compare (i) the system's performance and processing time with (ii) the scores and time spent by a human with more than 5 years of experience in recruitment. We focus on two main outcomes: processing time per candidate and the consistency between the system's recommendations and the human expert's decisions, e.g., to reject or move on to a technical interview.
Beyond this main question, the paper makes the following contribution:

\begin{enumerate}
\item Integrated multimodal design. We present a specific architecture for a web-based multimodal hiring assistant that combines resume analysis, semantic skill matching, red flag detection, public profile verification, and basic video analysis into a single pipeline driven by a large language model. The system is configurable and designed to be used by a technically competent hiring manager rather than a specialized data analytics team.
\item Time and quality evaluation protocol. We propose an evaluation scheme to compare AI-assisted candidate screening with human baselines that clearly captures both selection time and quality of the decision.
\item Insights on design and interpretability. Drawing on the system's implementation and pilot deployment, we discuss practical design decisions that help make AI recommendations more transparent and easier to override by humans.
\item Ethical and regulatory framework for screening based on public data. We describe how an AI-based validation system can use publicly available data while taking into account issues of privacy, bias, and regulatory requirements. In particular, we position the system as a decision support tool rather than an autonomous decision-maker, and highlight open issues regarding bias auditing, candidate consent, and explainability that need to be addressed in future work.
\end{enumerate}

\section{Related Work}

The idea of automating personnel hiring processes is not new. As early as 1983, Dave Bartram \cite{bartram1991micropat} proposed the MICROPAT automated system for the automated selection of aviation pilots. It was a rule-based personnel selection system that predated ATS. For Large Language Models (LLM), the trajectory effectively began with BERT: in 2019, one of the early papers to frame hiring tasks using transformer-based language models is work by Bhatia et al. \cite{bhatia2019resumebert}, which proposed a system that structured resumes, parsed the document, and used a BERT classifier to normalize non-standard text fragments. Since then, both the capabilities of such approaches and the breadth of the related research have expanded substantially. 

\subsection{Communications module}

The most standard component of hiring-process automation is communication with people. According to recent work by Wuttke et al. \cite{wuttke2024aiconvinterviewing}, in controlled experiments their LLM-based interviewer elicited more information from candidates than a human interviewer (52.39 words on average versus 32.81), and the ratings for relevance and clarity were higher by 0.3-0.4 points. On the other hand, this had a strong effect on respondents’ subjective evaluations: AI interviews were rated as less interesting (2.5 versus. 3.9) and associated with a lower willingness to repeat (2.5 versus. 3.6). This suggests that with careful prompting and improved human-likeness, it is possible to obtain higher-quality responses and increase scalability while maintaining respondent experience. In a study by S. Akram, P. Buono, and R. Lanzilotti \cite{akram2023recruitmentchatbot}, the authors also emphasize that, at the time of the study, chatbots most commonly struggled to demonstrate deep understanding, nuance, cultural sensitivity, and emotional support. However, given the rapid progress of LLMs and the fact that they are often evaluated primarily on how natural they sound rather than on practical utility, it would be valuable to conduct a comparable follow-up study. In any case, even then the conclusion was that chatbots are a useful tool for optimizing the process. 

\subsection{Resume evaluation}

Another task that was automated early on is resume evaluation. Initially, ATSs relied on rule-based approaches and keyword filters. These systems searched for exact matches of required skills or positions and assigned scores based on the frequency of the terms. But as models improved, current capabilities make it possible to move beyond exact wording and instead use semantic comparison to count candidates who describe equivalent experience using different terminology. For example, K. Khelkhal and D. Lanasri \cite{khelkhal2025smarthiring} applied heuristics to identify names, phone numbers, and years of experience, while for the skills list they applied the all-MiniLM-L6-v2 embedding model, which maps skill phrases into a shared vector space; they then computed cosine similarity and aggregated the results into a weighted overall score. At the same time, there are more technically advanced applications. For example, Lo et al. \cite{lo2025aihiringllms} developed a multi-agent LLM-based pipeline for resume review, where the key idea is to decompose the overall process into specialized agents and to use Retrieval-Augmented Generation (RAG) to incorporate external knowledge without additional fine-tuning. In this setup, the authors assessed skills, work experience, educational background, role alignment, and self-assessment. The resulting score structure is broadly similar to how human evaluations are distributed, while the authors provide a qualitative comparison against a single-LLM baseline and emphasize the benefit of RAG through a higher Pearson correlation. C. Gan, Q. Zhang, and T. Mori \cite{gan2024llmagentsrecruitment} investigated the quality of LLMs for resume assessment. Based on 50 manually annotated resumes, LLaMA2-13B achieves a ROUGE-1 score of 34.75 in a zero-shot setting, which increases to 37.30 after fine-tuning. GPT-4 showed a relatively high correlation, such that 11 of the 12 top-rated resumes matched the human assessment, while being approximately an order of magnitude faster. However, most existing resume scorers stop at producing a ranked list or a numeric score. They rarely provide explicit flags for suspicious career patterns or fine-grained justifications that can be easily inspected or adjusted by hiring managers.

\subsection{Technical assignment evaluation}

This topic is particularly interesting because it has largely progressed in a different direction. One of the core LLM use cases is assistance with code generation. Companies invest substantial effort into improving their models for this purpose or building specialized models, and there are also many datasets for evaluating LLM performance. One widely used benchmark is SWE-bench Verified \cite{openai2024swebenchverified}. However, in our context, it is necessary to consider systems that assess code written specifically by a human, i.e., approaches based on the LLM-as-a-Judge paradigm. For example, Jiang et al. \cite{jiang2025codejudgebench} examine the capabilities of 26 different models for judging tasks related to code generation, test generation, and code repair via pair-wise judging. A total of 5,352 examples were created using more capable LLMs, and Gemini-2.5-Pro performed best, achieving 82\% accuracy. The authors note that test generation is the most challenging task, and they also identify positional bias, which reaches up to 14\% in some models. Akyash et al. \cite{akyash2025stepgrade} proposed the StepGrade framework for automated assessment of programming assignments, where GPT-4 follows a sequence of reasoning steps via Chain-of-Thought (CoT) to evaluate code functionality, quality, and efficiency, and to produce both a score and detailed feedback. Compared to standard prompting, CoT yields more specific and actionable recommendations for improving the code. Its scores are generally closer to human ratings, as reflected by lower MAE per category.

\subsection{Visual evaluation}

Due to the annual ACM Multimedia (ACM MM) conference, a substantial number of improvements and new approaches for multimodal inputs emerge each year. For example, at ACM MM 2025 \cite{acmmm2025grandchallenges} topics such as multimodal sentiment, emotion and personality analysis, as well as multimedia verification were covered. As an illustration, within the conference context, Zhao et al. \cite{zhao2025depressionmllm} presented a framework for estimating a person’s level of depression. Based on Qwen2-Audio and a self-supervised vision encoder adapted via Parameter-Efficient Fine-Tuning (PEFT), the system achieves 0.825 F1 on the DAIC-WoZ test set, outperforming prior multimodal and unimodal analysis approaches. Regarding the assessment related to a specific interview, Laukaitis et al. \cite{laukaitis2025fairvid} propose the FAIR-VID framework for evaluating applicant statements (documents, completed forms, and video interviews), with an emphasis on transparency, human-in-the-loop operation, and an explicit note that the approach may be applicable to recruiting workflows. The system uses Gemma 3 27B as the multimodal model in a zero-shot setting and LayoutLMv3 Base as a layout-aware transformer operating over OCR-derived inputs. For video, language descriptions are generated for each frame, along with a JSON file containing predefined fields about the applicant and the surrounding context. The system performs offer prediction using the completed form and document, achieving 85\% precision and 82\% recall, while the additional use of video materials increased both metrics by 6 percentage points.

\subsection{Answer evaluation}

Answer assessment is a fairly common task where LLMs provide substantial support. As a result, recent papers often address adjacent issues. For example, Rao et al. \cite{rao2025invisiblefilters} examined how GPT-4o and Google Gemini 1.5 Flash evaluate interview transcripts across different cultural contexts. Even after swapping demographic attributes such as gender, caste, or region, a statistically significant difference remained between scores assigned to residents of India and those assigned to residents of the United Kingdom, and the authors attribute this gap to differences in linguistic features. Bergerhoff et al. \cite{bergerhoff2024autoconvassessment} studied the correlation between GPT-4 ratings and those of an assistant professor. The automated assessment module performed retrieval of relevant fragments from the course materials and autonomously generated follow-up questions and clarifications. Based on a survey of 21 students, the correlation with the examiner’s scores was 92\%. In addition, 82\% of students reported lower stress, and 61\% reported higher perceived fairness compared to a human examiner. On the other hand, 90\% of respondents believed the system could be gamed. A slightly different scenario was considered by Pack et al. \cite{pack2024llmaes}. They investigated the validity and reliability of a set of LLMs (PaLM 2, Claude 2, GPT-3.5, GPT-4) as automatic essay scoring systems for English language learners. Using 119 essays, they performed duplicate model scorings and compared them to human ratings, finding correlations in the 60-85\% range depending on the model and the scoring attempt. Interestingly, the OpenAI models showed lower correlation on the second scoring pass than on the first, whereas the other models exhibited improved similarity to human scores.

\subsection{Public data evaluation}

Entity resolution (ER) is a particularly active topic overall, largely because of the underlying optimization challenges. Information from the internet can be useful not only in the candidate selection domain, but in practice the volume of available data can be overwhelming. Zhang et al. \cite{zhang2025aiclentityresolution} propose using active in-context learning and domain-invariant similarity to improve context selection for LLMs and strengthen entity matching. The publicly available preview notes that improvements are observed across different LLMs. Fan et al. \cite{fan2023costeffectiveicl} developed the BATCHER framework for batch prompting, which reduces API costs without degrading quality. Using GPT-3.5 and GPT-4, their system achieves up to 7× cost savings while maintaining comparable quality. Munne et al. \cite{munne2025entityprofile} addressed the heterogeneity and incompleteness of data across different knowledge graphs. Their ProLEA framework proposes the following solution: GPT-4 generates entity profiles from attributes, relations, and context, then a BERT-based embedding module produces a shortlist of candidates using similarity and finally, GPT-4 performs re-ranking, applies a threshold to resolve conflicts, and provides feedback to improve the embedding model. Across multiple datasets, the approach reaches over 96\% quality, although disabling reasoning substantially reduces this performance. Cui et al. \cite{cui2025verifiablemisinfo} consider the problem from the perspective of verified misinformation detection, where the LLM not only classifies content but also reconstructs an evidence trail. Their agent, built as a multi-tool system, reports improved performance on three selected datasets (FakeNewsNet, LIAR, and COVID-19) compared to a baseline GPT-4o (for example, F1 on FakeNewsNet is 89.3 for the proposed agent versus 84.7 for standalone GPT-4o). Despite their practical appeal, publicly-based screening is controversial in research and ethics discussions. Concerns include privacy and proportionality, potential bias and opacity. Given that, our system analyzes data that is only publicly accessible and primarily uses to verify key claims rather than to drive the main scoring.

\subsection{Culture fit evaluation}

While attractive to practitioners, this area is scientifically and ethically fraught. Validity of inferred traits is often unclear, training data may encode existing biases, and over-reliance on culture fit raises the risk of homogeneity and discrimination against candidates with different backgrounds or communication styles. In Hoffmann et al. \cite{hoffmann2025llmbehaviorhiring}, the authors propose an econometric framework to identify which signals LLMs implicitly weight, whether scoring logic differs across demographic groups, and how these patterns can be compared to human preferences. They find that, in general, the model’s behavior aligns with intuitive signals derived from similar information. However, the same attributes may be evaluated under different rules for different demographic profiles. Kim et al. \cite{kim2025blindedbycontext} examine the halo effect in hiring, where an overall impression shifts ratings across multiple competencies, and show that a similar phenomenon can also arise in MLLM systems when they are influenced by irrelevant multimodal context. Li et al. \cite{li2024culturellm} evaluate cultural bias driven by the dominance of English-language corpora in GPT-3.5, contrasting it with custom CultureLLM models.

\subsection{Integrated multi-modal hiring assistants}

There is a growing body of research on integrated, multi-modal AI hiring assistants. Some recent academic systems combine modules for resume parsing, semantic matching, and basic interview analysis into unified pipelines, sometimes orchestrated via multi-agent LLM frameworks. Industrial platforms increasingly advertise "end-to-end AI recruiting," bundling resume scoring, chatbots, scheduling, and analytics into a single product.

However, even in this integrated space, several gaps remain:

\begin{itemize}
\item Modularity and configurability. Many platforms expose limited control over how signals from different sources are weighted or combined. This makes them hard to adapt to the needs of a small, technically sophisticated company that wants to tune behavior per role and retain control over decision logic.
\item Transparent reasoning. A substantial portion of existing systems behave like black boxes, they output scores or recommendations without exposing intermediate reasoning steps.
\item Explicit evaluation of time-quality trade-offs. While vendors frequently claim large time savings, there are relatively few detailed, peer-reviewed evaluations that quantify both processing time and agreement with experienced human recruiters on the same dataset.
\end{itemize}

Our system sits in this integrated multi-modal category but aims to address these gaps explicitly. It is conceptually aligned with these LLM-based approaches but organizes candidate validation as a deterministic pipeline of interpretable modules orchestrated by an LLM, exposes configuration in a human-editable way and is evaluated not only for decision quality but also for the practical outcome that matters most in our context: how much time it saves an experienced hiring professional while maintaining comparable screening quality. This design emphasizes configurability and transparency for a technically competent hiring manager, rather than relying on an unclear end-to-end neural model.

\section{Methodology}

The proposed system is a modular, multi-agent pipeline that initially takes as input a candidate’s resume, a technical assignment, and answers to questions provided in either text or video format, and then performs step-by-step processing. A schematic overview is shown in \hyperref[fig:diagram]{Fig~\ref*{fig:diagram}}. First, the inputs are collected and harmonized into a Markdown representation. Next, the context-construction module uses an agent to infer the candidate’s context in terms of skills and culture and compares it against the job description and the company’s values. All available information is then consolidated within the public-data analysis module, which reviews all permissible information about the candidate, verifies it, and checks it for consistency. This enables a comprehensive view of the candidate’s profile and helps identify missing points, which are subsequently addressed by a dedicated communications module that requests specific clarifications, questions or links. The resulting context is aggregated and passed to the next agent, which assesses the candidate’s skills and overall fit with the company’s requirements by computing a corresponding score. All candidates are then assembled into a ranking using custom weights for each candidate aspect, together with a detailed report, and the top 10 candidates selected by the system are forwarded to the recruiter for validation. With a report that transparently explains any score reductions and highlights the candidate’s strengths and weaknesses, the recruiter can quickly review the profile, verify key claims, and make a final decision.

\begin{figure*}
    \centering
    \includegraphics[width=\textwidth, height=0.95\textheight, keepaspectratio]{chart_diagram}
    \caption{Schematic overview of the proposed modular multi-agent pipeline for candidate data ingestion, context construction, public-data verification, scoring, ranking, and validation.}
    \label{fig:diagram}
\end{figure*}

\subsection{Preprocessing}

Typically, candidate filtering begins once a company publishes a vacancy on a job-posting platform. This may be a specialized website (e.g., Indeed) or the company’s own website. This leads candidates to apply and submit their resumes, which are ingested by the first preprocessing module. The core engine is the GPT-4o LLM accessed through the OpenAI API and constrained by a strict, normalized JSON schema, which makes it possible to determine what the candidate has submitted and what is missing. A large share of candidates indicate that they want to work but forget to submit even a resume, so an automated agent will communicate politely with the candidate until it receives the resume, the technical assignment solution files (depending on the role), and answers to a short set of behavioral questions, which are also defined depending on the role. Video responses are preferred, but based on our experience, on average 4 out of 5 candidates ultimately do not submit a video. Therefore, we allow text responses, while warning that the candidate’s score will be slightly reduced. Next, file processing is performed: for resumes, information is extracted using PaddleOCR for optical character recognition, which prevents prompt-injection via hidden instructions embedded in the resume. For the technical assignment, depending on the task, an efficiency score is computed that accounts for speed and quality on dedicated test sets with diverse edge cases, along with an LLM-based assessment under predefined parameters of the codebase characteristics (format consistency, type checking, complexity, duplication, presence of tests, etc.), quality evaluation, and risks related to security and code copying. For video analysis, the ffmpeg library is used to sample 1 frame every 5 seconds and to extract the audio track. Images are described by GPT-4o Vision to provide an overall assessment of candidate professionalism using explicitly defined characteristics, such as acceptable attire, non-excessive makeup, the face being visible on video, and clearly inappropriate or concerning symbols in the background. Our system adopts a conservative stance on video analysis, rather than attempting to infer personality traits or deep psychological characteristics, it uses video inputs to verify that the candidate is the same person across documents and profiles, capture obvious visual/contextual issues and enrich the qualitative understanding of communication style. The candidate’s photo from the resume (if available) is also provided to the same model to check similarity and to complement the professionalism assessment. For transcription, the OpenAI Whisper speech-to-text model is used. All of these outputs, converted into textual descriptions, are harmonized and transformed into Markdown using the Docling library.

\subsection{Initial Context}

All data are then passed to the base-context generation module. Using the job description and company information, a GPT-4o-based agent, constrained by a strict normalized JSON schema, is enforced to extract the candidate’s personal data, including education, employment history, skills, projects, technologies, languages, achievements, and related attributes, capturing useful information not only from the corresponding resume sections but also across the materials as a whole. Skills are normalized via an alias map and, when needed, are expanded through a graph or ontology. Through careful prompting, a low temperature setting, and a JSON-only configuration, we obtained stable initial candidate profiles with minimal variability.

\subsection{Context Augmentation}

The next GPT-4o agent uses Model Context Protocol (MCP) tooling and the APIs of specific services to access the internet and to retrieve and analyze candidate information from LinkedIn, GitHub, and other social platforms such as Meta, Instagram, WhatsApp, Viber, Telegram, and X, as well as to search for broader public traces with additional use of DuckDuckGo. For the accounts that are identified, the agent collects and validates available information about employment, skills, and other relevant details to confirm that the biography is consistent, minimizing the risk of linking to incorrect accounts while increasing the evidentiary threshold for key facts. The collected data are then used for a consistency check: extracting similarities, discrepancies, and red flags, including indicators of deception, calls to terrorism, or manifestations of toxicity. If the agent has substantial uncertainty and such links are not provided in the resume, it contacts the candidate through the communications module and requests the additional information. As a result, a complete candidate profile is produced, which also accounts for the candidate’s credibility and online presence.

\subsection{LLM Analysis}

The next GPT-4o-based agent uses the full candidate profile for evaluation. First, Named Entity Recognition (NER) is applied to extract all explicit mentions of technologies and relevant attributes. Next, Entity Disambiguation (ED) resolves ambiguities by separating specific entity mentions from background context - for example, "react," which may be used either as a verb or as a framework. Each entity is then mapped via Named Entity Linking (NEL), producing a structured representation that connects each entity to the skills list, achievements, attributes, and related categories. This makes it possible to distinguish general candidate statements from supporting evidence with an associated confidence level. This structure is then used for evaluation that combines semantic matching with LLM-based matching, including evidence scoring based on direct experience, projects, answers to questions, and GitHub activity. A separate depth check is also performed to identify indicators of genuine understanding, such as awareness of limitations, common pitfalls, metrics, and trade-offs. In addition, the agent searches for predefined negative indicators, including vague phrasing, date inconsistencies, long employment gaps, multiple concurrent jobs, signs of AI-generated content, incomplete profiles, and a lack of publicly available supporting evidence. All of these steps are implemented using GPT-4o with a strict normalized JSON configuration and a multi-step prompting sequence to ensure comprehensive use of the available context. The module outputs are aggregated into a compact set of interpretable scores with explanations for technical fit, culture fit, and risks. Technical fit reflects how well the candidate matches the project and role requirements, while culture fit evaluates not only similarity but also the candidate’s general tendencies in problem-solving, collaboration, communication, leadership signals, and initiative across seven predefined categories (work style, collaboration, communication, growth mindset, ownership, innovation, and a values list) according to the text description of the company and project values. The goal is not to predict personality in a psychometric sense, but to help the hiring manager quickly see whether there is any observable support in the candidate’s materials for the values the company claims to care about.

\subsection{Candidates ranking}

The technical fit and culture fit scores are already aggregated indicators of a candidate’s similarity to the role, as determined by the previous module. All risk penalties are reformatted onto a 0-1 scale. Accordingly, the formula for computing a single numeric score is:
\begin{equation}
S_{\text{total}}=\beta \cdot T_{\text{tech}} + (1-\beta)\cdot T_{\text{culture}} - R,
\label{eq:total_score}
\end{equation}
where $S_{\text{total}}$ is the final numeric candidate score used for ranking (higher is better), $T_{\text{tech}}\in[0,1]$ is the aggregated technical fit score produced by the previous module, $T_{\text{culture}}\in[0,1]$ is the aggregated culture fit score produced by the previous module, $R\in[0,1]$ is the aggregated risk penalty normalized to a $0$-$1$ scale (higher values correspond to higher risk and therefore a stronger reduction of the final score), and $\beta\in[0,1]$ is a weighting coefficient that controls the trade-off between technical and cultural fit ($\beta \to 1$ emphasizes $T_{\text{tech}}$, while $\beta \to 0$ emphasizes $T_{\text{culture}}$). The module also generates a concise report outlining each candidate’s strengths and weaknesses, as well as the reasons for any score reductions. The profiles of the 10 candidates with the highest scores are forwarded to the next stage for validation.

\subsection{Human-in-the-loop validation }

A professional recruiter, using the reports via a Gradio interface, can efficiently review candidates, verify claims, and make a final decision for each one. The human reviewer can also request the next batch of candidates and provide feedback. Claude Code powered by Opus 4.5 has access to the full repository and detailed documentation, which enables rapid incorporation of feedback and immediate verification of correctness using a broad test suite. 

\section{Experimental results}

Our AI-driven candidate validation system is implemented in Python and relies on open-source libraries for data preprocessing and orchestration. Document harmonization into Markdown is performed with Docling; video preprocessing uses ffmpeg for frame sampling and audio extraction; and PaddleOCR is used to parse text from scanned PDFs, serving also as an additional safeguard against hidden-instruction prompt injection. The human-in-the-loop validation interface is implemented in Gradio. The LLM and speech-to-text components are integrated via the OpenAI API (GPT-4o, text-embedding-3-small, Whisper), with critical stages - structured extraction, NER/ED/NEL, evidence scoring, and final aggregation - executed in a strict JSON-only mode with a controlled multi-step prompting sequence to minimize output variability. For context augmentation, the system accesses public sources through MCP-based tools and performs additional search via DuckDuckGo. The pipeline runs on hardware equipped with an Nvidia GeForce RTX 3050, enabling efficient end-to-end processing. When local deployment is required under stricter privacy constraints, the system can be adapted to local LLM runtimes such as Ollama while preserving the overall architecture.  

\subsection{Dataset}

We evaluated the system on a dataset of 64 resumes of middle-level Python backend engineers. The dataset was independently reviewed by two human baselines: a professional recruiter with more than five years of experience and a recruiter with approximately one year of experience. Each recruiter marked the candidates whom they would advance to a technical interview for this role. Using the decisions of the more experienced recruiter as a reference, we can examine how well the system reproduces the boundary between candidates who are considered strong enough to proceed to a technical interview and those who should be filtered out at the validation stage, and we can also compare its suggestions with the choices made by the less experienced recruiter. At the same time, it is important to acknowledge that this design inherits any biases present in the recruiters’ judgments, and that the labels reflect decisions for a specific position and seniority level rather than a universal ground truth.

\subsection{Evaluation metrics}

For evaluating the candidate validation system we treat the decisions of the more experienced recruiter as the reference classification and compute precision and recall of the system with respect to this reference. To capture both the speed of operation and the implicit cost of false positives and false negatives, we propose an aggregate efficiency metric based on the expected time spent per genuinely suitable candidate. Formally, we define  
\begin{equation}
T_{\mathrm{avg}}=\frac{T_{\mathrm{scr}}}{qR}+\frac{T_{\mathrm{tech}}}{P},
\end{equation}
where \(T_{\mathrm{avg}}\) denotes the average total time required to validate and technically interview one truly suitable candidate, \(T_{\mathrm{scr}}\) is the average time in hours needed to screen a single candidate at the validation stage, \(T_{\mathrm{tech}}\) is the average duration in hours of one technical interview, \(R\) is the recall of the system with respect to the professional recruiter, \(P\) is the corresponding precision and \(q\) captures the base
rate of suitable candidates in the applicant pool. When \(P = 1\) and \(R = 1\), the expression reduces to 
\begin{equation}
T_{\mathrm{avg}} = \frac{T_{\mathrm{scr}}}{q} + T_{\mathrm{tech}},\
\end{equation}
which corresponds to spending required number of screening passes and only one technical interview per good candidate. Lower recall increases the first term and reflects the fact that more candidates must be screened to identify the same number of good ones; lower precision increases the second term and reflects additional technical interviews spent on candidates who ultimately turn out not to be suitable. In this way \(T_{\mathrm{avg}}\) approximates the average time cost of obtaining one good candidate and discourages strategies in which a recruiter or system quickly passes a large fraction of applicants to technical interview, thereby shifting the burden onto technical specialists. If desired, the metric can be extended with weights to reflect different monetary or opportunity costs of recruiter and engineer time, for example  
\begin{equation}
T_{\mathrm{avg}}^{(w)} = w_{\mathrm{scr}}\frac{T_{\mathrm{scr}}}{qR} + w_{\mathrm{tech}}\frac{T_{\mathrm{tech}}}{P},
\end{equation}
where \(w_{\mathrm{scr}}\) and \(w_{\mathrm{tech}}\) are positive coefficients that encode the relative importance or cost of time spent at the screening and technical-interview stages.

\subsection{Performance}

\begin{figure}[t]
    \centering
    \includegraphics[width=\linewidth]{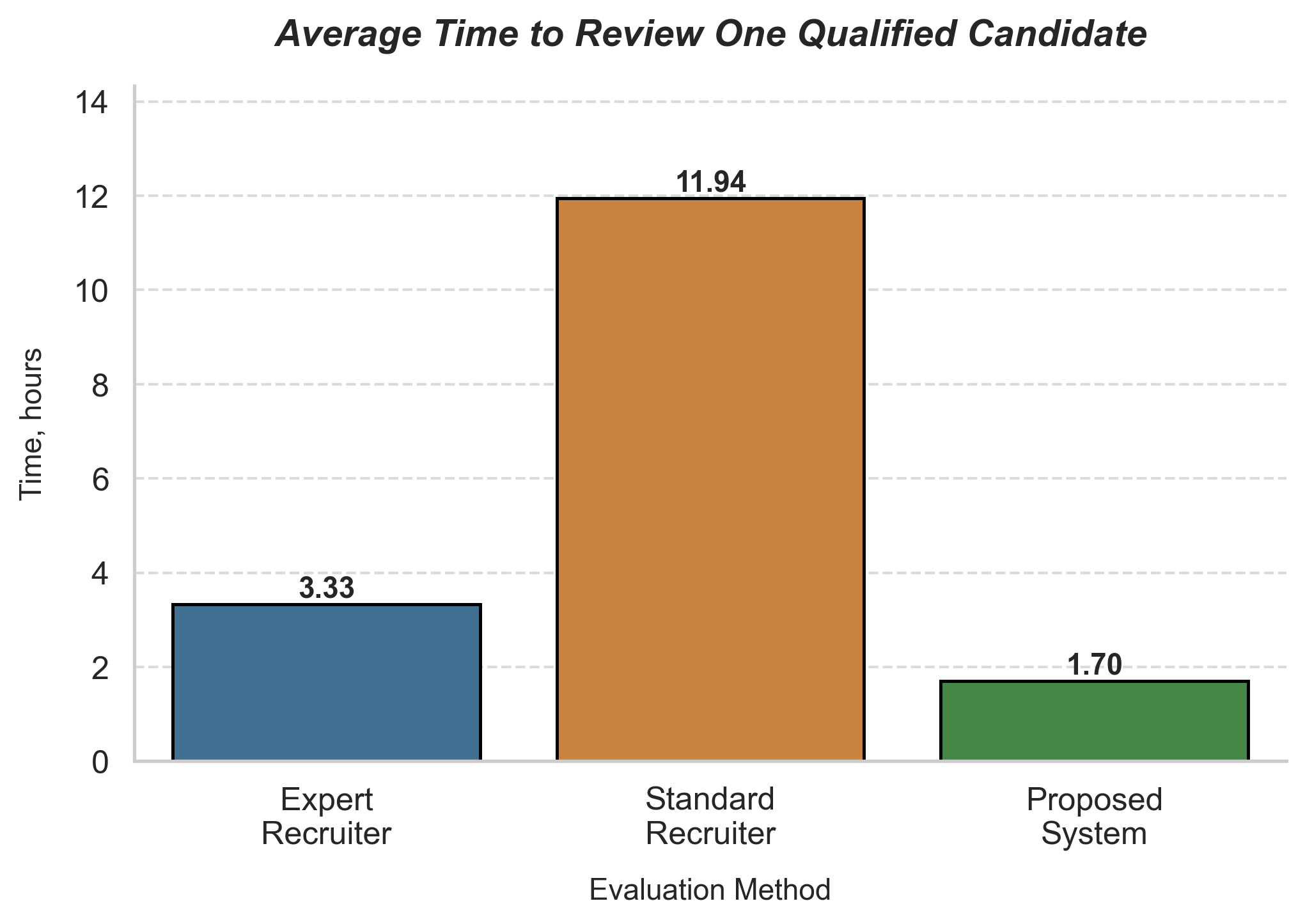}
    \caption{Comparison of average review time per qualified candidate for expert recruiters, standard recruiters, and the proposed system.}
    \label{fig:average_time}
\end{figure}

\begin{figure}[t]
    \centering
    \includegraphics[width=\linewidth]{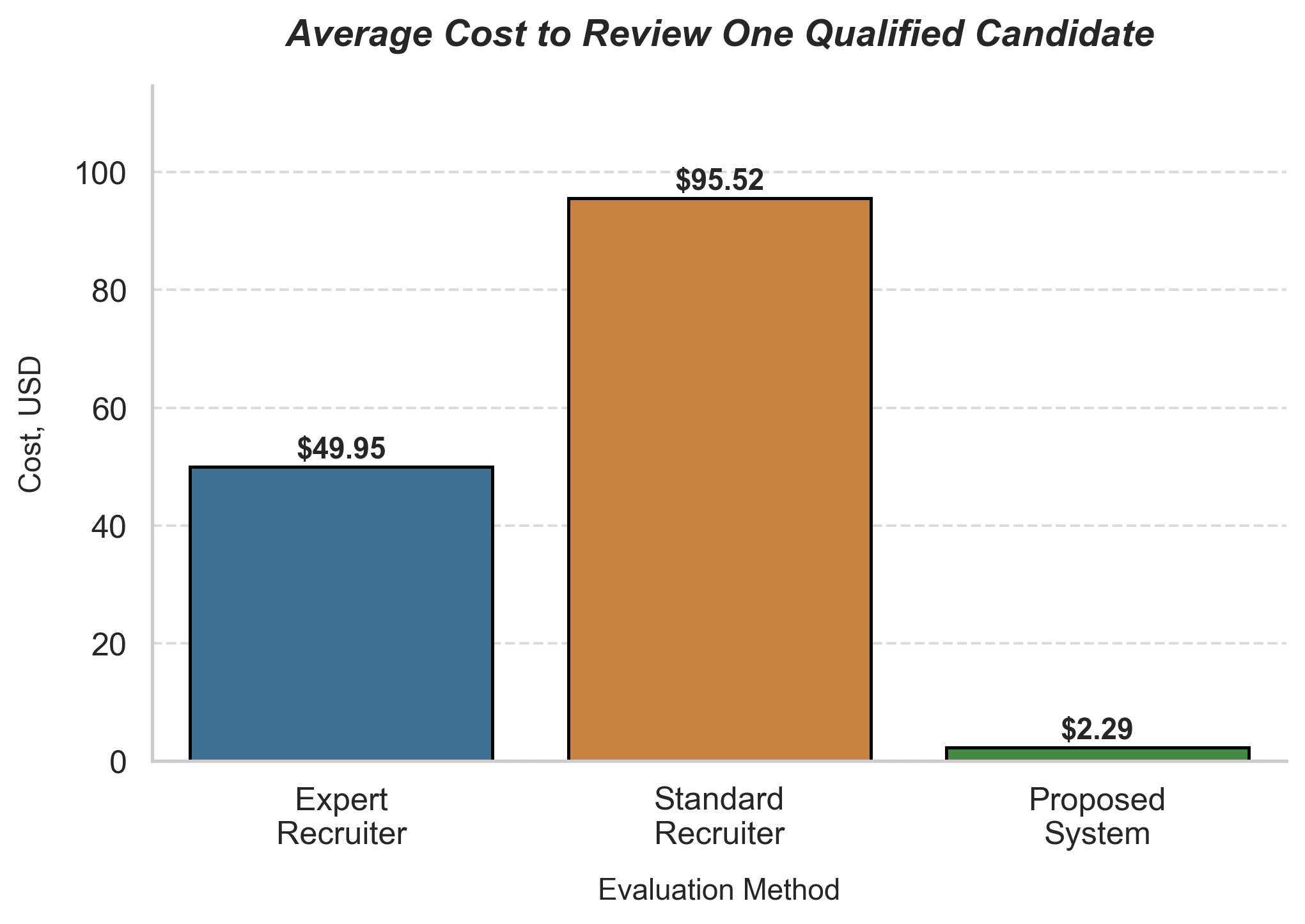}
    \caption{Comparison of average review cost per qualified candidate for expert recruiters, standard recruiters, and the proposed system.}
    \label{fig:average_cost}
\end{figure}

To make the error structure explicit, we additionally report confusion matrices against the professional recruiter as a reference: the less experienced recruiter vs. professional labels are shown in \hyperref[tab:cm_novice_vs_pro]{Table~\ref*{tab:cm_novice_vs_pro}}, while the proposed system vs. professional labels are shown in \hyperref[tab:cm_system_vs_pro]{Table~\ref*{tab:cm_system_vs_pro}}. On average, the professional recruiter in our company reviews approximately 1.07 candidates per hour and paid approximately \$15 per hour, whereas the less experienced recruiter in our sample processes about 0.33 candidates per hour with a salary of \$8 per hour. A substantial share of time is spent on candidate communication, verifying information using public data sources, and repeatedly switching between processes. In comparison, over the same period our system processes approximately 3.28 candidates. Taking into account the observed precision and recall metrics and assuming an average technical interview duration of 0.5 hours and approximately one suitable candidate per 3 applicants, the resulting average time cost per truly suitable candidate, as defined by the efficiency measure in the previous subsection, is shown in \hyperref[fig:average_time]{Fig~\ref*{fig:average_time}} for both human baselines and for the system. Under this metric the system achieves an average of 1.70 hours per good candidate, which is better than the time spent by a professional recruiter 3.33 hours and substantially better than the less experienced recruiter. This outcome is driven by a combination of a relatively low error rate and a significantly higher validation throughput compared to individual human screeners. When the financial aspect is considered, the gap becomes even more pronounced: in addition to compute, we account for OpenAI API usage \cite{openai_pricing_docs} for GPT-4o text processing (\$2.50 per 1M input and \$10.00 per 1M output tokens), text-embedding-3-small embeddings (\$0.02 per 1M tokens), Whisper speech-to-text transcription (\$0.006 per minute) and GPT-4o Vision (\$0.0028 per one frame with a resolution 1920×1080, since the price for 1280×720 will be the same). Under this accounting, the estimated average system processing cost is \$2.29 per qualified candidate versus \$50 for an expert recruiter, as shown in  \hyperref[fig:average_cost]{Fig~\ref*{fig:average_cost}}. For reference, the cost for a typical recruiter is approximately \$95 per qualified candidate. 

\begin{table}[t]
\caption{Confusion matrix for the less experienced recruiter against the professional recruiter (professional labels as reference).}
\label{tab:cm_novice_vs_pro}
\centering
\setlength{\tabcolsep}{6pt}
\renewcommand{\arraystretch}{1.15}
\begin{tabular}{lcc}
\hline
 & \multicolumn{2}{c}{Predicted (Less experienced recruiter)} \\
\cline{2-3}
Actual (Professional) & Qualified (+) & Not qualified (--) \\
\hline
Qualified (+)      &  
                    14&  
                    7\\
Not qualified (--) &  
                    3&  
                    40\\
\hline
\end{tabular}
\end{table}

\begin{table}[t]
\caption{Confusion matrix for the proposed system against the professional recruiter (professional labels as reference).}
\label{tab:cm_system_vs_pro}
\centering
\setlength{\tabcolsep}{6pt}
\renewcommand{\arraystretch}{1.15}
\begin{tabular}{lcc}
\hline
 & \multicolumn{2}{c}{Predicted (System)} \\
\cline{2-3}
Actual (Professional) & Qualified (+) & Not qualified (--) \\
\hline
Qualified (+)      &  
                    16&  
                    5\\
Not qualified (--) &  
                    2&  
                    41\\
\hline
\end{tabular}
\end{table}


\section{Discussion}

With the growing maturity of multi-agent architectures, such systems are increasingly being applied across diverse domains to support complex decision-making processes. In this context, our system demonstrates a measurable improvement over less experienced recruiters in the early-stage validation of candidates. Importantly, it does not only output a single numeric score but also produces a detailed, modular report with transparent explanations of each component score. This interpretability makes the system potentially useful not only as an automated screener but also as a decision-support tool for professional recruiters, by accelerating their review of candidates and providing all available information in a single, structured view that can then be validated and finalized by a human. As discussed earlier, the architecture can also be deployed in a fully local configuration without transmitting personal data to third parties, which is essential in privacy-sensitive environments.

\subsection{Conclusion \& Future Recommendations}

In this work, we proposed a multimodal approach to candidate validation that jointly considers resumes, short screening responses and publicly available information in order to assess technical skills, culture fit and salient risk factors. The framework is implemented as a pipeline of five main modules, which produce not only aggregate numeric scores but also a fully traceable report with explanations for each component. By combining standard computational methods with large language models, the system is designed to be robust to variations in wording and formatting and to operate across different roles without relying on hand-crafted features for a single vacancy. To evaluate its effectiveness, we tested the system on real candidates and compared its recommendations and efficiency against recruiters with different levels of experience. The results indicate that the proposed approach outperforms a professional recruiter, highlighting its usefulness as both an automation mechanism and a scalable AI-based decision-support solution in hiring contexts.

There remains substantial untapped potential in working with multimodal candidate data. Incorporating more powerful foundation models and advanced reasoning methods with expanded MCP instruments could further improve decision quality, even if such enhancements may introduce additional latency. Beyond the validation stage, there is scope for automating adjacent parts of the pipeline, such as drafting and publishing job postings, organizing screening interviews, and generating follow-up questions tailored to individual candidates. At the same time, the present implementation warrants a more extensive empirical evaluation. Future studies should consider a larger and more heterogeneous set of resumes, spanning multiple roles and seniority levels, and compare the system’s performance against a broader panel of recruiters. To reduce individual rater bias and enable statistically grounded conclusions, each candidate should be independently evaluated at least 5 experienced recruiters, blinded to the system output and guided by a shared scoring rubric. The aggregated panel score can then serve as a reference while inter-rater reliability and paired bootstrap or permutation tests can be used to quantify uncertainty and statistical significance. As with any socio-technical system, both humans and models introduce biases, it is therefore necessary to systematically investigate whether the system treats candidates of different genders, ethnicities and backgrounds fairly, and to identify where its weaknesses lie. This also raises the question of how the system evaluates artificially generated or heavily model-assisted resumes, which might be systematically advantaged if their style aligns closely with the evaluation model. Finally, there is an important ethical question of transparency in deployment: in real hiring workflows, AI-generated recommendations should be clearly indicated as such, so that their role in the decision process is visible and accountable.

\vspace{12pt}


\begin{thebibliography}{99}

\bibitem{bartram1991micropat}
D.~Bartram and H.~C.~A.~Dale, ``Validation of the micropat battery of pilot aptitude tests,'' in \emph{Advances in Computer-Based Human Assessment}. Dordrecht, The Netherlands: Springer, 1991, pp.~149--169, doi: 10.1007/978-94-011-3322-7\_5.

\bibitem{bhatia2019resumebert}
V.~Bhatia, P.~Rawat, A.~Kumar, and R.~R.~Shah, ``End-to-end resume parsing and finding candidates for a job description using {BERT},'' \emph{arXiv preprint} arXiv:1910.03089, 2019, doi: 10.48550/arXiv.1910.03089. [Online]. Available: \url{https://arxiv.org/abs/1910.03089}

\bibitem{wuttke2024aiconvinterviewing}
A.~Wuttke \emph{et al.}, ``{AI} conversational interviewing: Transforming surveys with {LLM}s as adaptive interviewers,'' \emph{arXiv preprint} arXiv:2410.01824, 2024, doi: 10.48550/arXiv.2410.01824. [Online]. Available: \url{https://arxiv.org/abs/2410.01824}

\bibitem{akram2023recruitmentchatbot}
S.~Akram, P.~Buono, and R.~Lanzilotti, ``Recruitment chatbot acceptance in company practices: An elicitation study,'' in \emph{Proc. ACM SIGCHI Italian Chapter Conf. (CHItaly)}, 2023, doi: 10.1145/3605390.3605420. [Online]. Available: \url{https://www.researchgate.net/publication/374054649_Recruitment_Chatbot_Acceptance_in_Company_Practices_An_Elicitation_Study}

\bibitem{khelkhal2025smarthiring}
K.~Khelkhal and D.~Lanasri, ``Smart-hiring: An explainable end-to-end pipeline for {CV} information extraction and job matching,'' \emph{arXiv preprint} arXiv:2511.02537, 2025, doi: 10.48550/arXiv.2511.02537. [Online]. Available: \url{https://arxiv.org/abs/2511.02537}

\bibitem{lo2025aihiringllms}
F.~P.-W.~Lo \emph{et al.}, ``{AI} hiring with {LLM}s: A context-aware and explainable multi-agent framework for resume screening,'' \emph{arXiv preprint} arXiv:2504.02870, 2025, doi: 10.48550/arXiv.2504.02870. [Online]. Available: \url{https://arxiv.org/abs/2504.02870}

\bibitem{gan2024llmagentsrecruitment}
C.~Gan, Q.~Zhang, and T.~Mori, ``Application of {LLM} agents in recruitment: A novel framework for automated resume screening,'' \emph{Journal of Information Processing}, vol.~32, pp.~881-893, 2024, doi: 10.2197/ipsjjip.32.881.

\bibitem{openai2024swebenchverified}
OpenAI, ``Introducing {SWE}-bench Verified,'' Aug.~13, 2024 (updated Feb.~24, 2025). [Online]. Available: \url{https://openai.com/index/introducing-swe-bench-verified/}

\bibitem{jiang2025codejudgebench}
H.~Jiang, Y.~Chen, Y.~Cao, H.-y.~Lee, and R.~T.~Tan, ``CodeJudgeBench: Benchmarking {LLM}-as-a-judge for coding tasks,'' \emph{arXiv preprint} arXiv:2507.10535, 2025, doi: 10.48550/arXiv.2507.10535. [Online]. Available: \url{https://arxiv.org/abs/2507.10535}

\bibitem{akyash2025stepgrade}
M.~Akyash, K.~Zamiri~Azar, and H.~Mardani~Kamali, ``StepGrade: Grading programming assignments with context-aware {LLM}s,'' \emph{arXiv preprint} arXiv:2503.20851, 2025, doi: 10.48550/arXiv.2503.20851. [Online]. Available: \url{https://arxiv.org/abs/2503.20851}

\bibitem{acmmm2025grandchallenges}
ACM Multimedia 2025, ``Grand Challenges,'' 2025. [Online]. Available: \url{https://acmmm2025.org/grand-challenge/}

\bibitem{zhao2025depressionmllm}
X.~Zhao \emph{et al.}, ``It hears, it sees too: Multi-modal {LLM} for depression detection by integrating visual understanding into audio language models,'' \emph{arXiv preprint} arXiv:2511.19877, 2025, doi: 10.48550/arXiv.2511.19877. [Online]. Available: \url{https://arxiv.org/abs/2511.19877}

\bibitem{laukaitis2025fairvid}
A.~Laukaitis \emph{et al.}, ``FAIR-VID: A multimodal pre-processing pipeline for student application analysis,'' \emph{Applied Sciences}, vol.~15, no.~24, Art. no.~13127, 2025, doi: 10.3390/app152413127.

\bibitem{rao2025invisiblefilters}
P.~S.~B.~Rao, L.~N.~Venkatesan, M.~Cherubini, and D.~B.~Jayagopi, ``Invisible filters: Cultural bias in hiring evaluations using large language models,'' \emph{arXiv preprint} arXiv:2508.16673, 2025, doi: 10.48550/arXiv.2508.16673. [Online]. Available: \url{https://arxiv.org/abs/2508.16673}

\bibitem{bergerhoff2024autoconvassessment}
J.~Bergerhoff \emph{et al.}, ``Automatic conversational assessment using large language model technology,'' in \emph{Proc. 2024 16th Int. Conf. on Education Technology and Computers}, 2024, doi: 10.1145/3702163.3702169.

\bibitem{pack2024llmaes}
A.~Pack, A.~Barrett, and J.~Escalante, ``Large language models and automated essay scoring of English language learner writing: Insights into validity and reliability,'' \emph{Computers and Education: Artificial Intelligence}, vol.~6, Art. no.~100234, 2024, doi: 10.1016/j.caeai.2024.100234.

\bibitem{zhang2025aiclentityresolution}
Z.~Zhang, W.~Zeng, J.~Tang, H.~Huang, and X.~Zhao, ``Active in-context learning for cross-domain entity resolution,'' \emph{Information Fusion}, vol.~117, Art. no.~102816, 2025, doi: 10.1016/j.inffus.2024.102816.

\bibitem{fan2023costeffectiveicl}
M.~Fan \emph{et al.}, ``Cost-effective in-context learning for entity resolution: A design space exploration,'' \emph{arXiv preprint} arXiv:2312.03987, 2023, doi: 10.48550/arXiv.2312.03987. [Online]. Available: \url{https://arxiv.org/abs/2312.03987}

\bibitem{munne2025entityprofile}
R.~F.~Munne, M.~M.~Rahman, and Y.~Matsumoto, ``Entity profile generation and reasoning with {LLM}s for entity alignment,'' in \emph{Findings of the Association for Computational Linguistics: EMNLP 2025}, Suzhou, China, 2025, pp.~20073--20086, doi: 10.18653/v1/2025.findings-emnlp.1093. [Online]. Available: \url{https://aclanthology.org/2025.findings-emnlp.1093/}

\bibitem{cui2025verifiablemisinfo}
Z.~Cui, T.~Huang, C.-E.~Chiang, and C.~Du, ``Toward verifiable misinformation detection: A multi-tool {LLM} agent framework,'' \emph{arXiv preprint} arXiv:2508.03092, 2025, doi: 10.48550/arXiv.2508.03092. [Online]. Available: \url{https://arxiv.org/abs/2508.03092}

\bibitem{hoffmann2025llmbehaviorhiring}
M.~Hoffmann, E.~Jouffroy, W.~Jouanneau, M.~Palyart, and C.~Pebereau, ``Evaluating {LLM} behavior in hiring: Implicit weights, fairness across groups, and alignment with human preferences,'' in \emph{Proc. 5th Workshop on Recommender Systems for Human Resources (RecSys-in-HR 2025)}, CEUR Workshop Proc., vol.~4046, 2025. [Online]. Available: \url{https://ceur-ws.org/Vol-4046/RecSysHR2025-paper_4.pdf}

\bibitem{kim2025blindedbycontext}
K.~Kim, J.~Ryu, H.~Jeon, and B.~Suh, ``Blinded by context: Unveiling the halo effect of {MLLM} in {AI} hiring,'' in \emph{Findings of the Association for Computational Linguistics: ACL 2025}, Vienna, Austria, 2025, pp.~26067--26113, doi: 10.18653/v1/2025.findings-acl.1338. [Online]. Available: \url{https://aclanthology.org/2025.findings-acl.1338/}

\bibitem{li2024culturellm}
C.~Li, M.~Chen, J.~Wang, S.~Sitaram, and X.~Xie, ``CultureLLM: Incorporating cultural differences into large language models,'' \emph{arXiv preprint} arXiv:2402.10946, 2024, doi: 10.48550/arXiv.2402.10946. [Online]. Available: \url{https://arxiv.org/abs/2402.10946}

\bibitem{openai_pricing_docs}
OpenAI, ``Pricing,'' \emph{OpenAI API Documentation}. [Online]. Available: \url{https://platform.openai.com/docs/pricing}. Accessed: Dec. 16, 2025.

\end{thebibliography}
\end{document}